\documentstyle[aclap]{article}

\long\def\comment#1{}
\newcommand{\tuple}[1]{\mbox{$\langle #1 \rangle$}}

\newcommand{\smes}{{\sc smes}}
\newcommand{\mona}{{\sc mona}}
\newcommand{\morphix}{{\sc morphix}}
\newcommand{\cosma}{{\sc cosma}}

\title{An Information Extraction Core System 
       for \\  Real World German Text Processing}
\author{G\"unter Neumann\thanks{
DFKI~GmbH,
Stuhlsatzenhausweg 3,
66123 Saarbr\"ucken, Germany,
{\tt neumann@dfki.uni-sb.de}} \And 
Rolf Backofen\thanks{
LMU,
Oettingenstrasse 67,
80538 M\"unchen, Germany,
{\tt backofen@informatik.uni-muenchen.de}} \And 
Judith Baur\thanks{
DFKI GmbH,
{\tt baur@dfki.uni-sb.de}} \And 
Markus Becker\thanks{
DFKI GmbH,
{\tt mbecker@dfki.uni-sb.de}} \And 
Christian Braun\thanks{
DFKI GmbH,
{\tt cbraun@dfki.uni-sb.de}}
}

\begin{document}

\maketitle

\begin{abstract}
This paper describes SMES, an information extraction core system
for real world German text processing. The basic design criterion of the 
system is of providing a set of basic powerful, robust, 
and efficient natural language components and generic linguistic 
knowledge sources which can easily be customized for processing 
different tasks in a  flexible manner. 
\end{abstract}

\section{Introduction}
\label{introduction}

There is no doubt that the amount of textual information electronically
available today has passed its critical mass leading to the emerging problem
that the more electronic text data is available the more difficult it is
to find or extract relevant information.
In order to overcome this problem new technologies for future
information management systems are explored by various
researchers. One new line of such research is the investigation and
development of {\em information extraction} (IE) systems.
The goal of IE is to build systems that find and link relevant information from
text data while ignoring extraneous and irrelevant information
\cite{CowieLehnert:96}. 

Current IE systems are to be quite
successfully in automatically processing large text collections with
high speed and robustness (see \cite{muc-6},
\cite{Chinchoretal:93}, and \cite{GrishmanSundheim:96b}). 
This is due to the fact that they can
provide a partial understanding of specific types of text with a certain 
degree of partial accuracy using
fast and robust shallow processing strategies (basically finite state 
technology). They have been ``made sensitive'' to certain
key pieces of information and thereby  provide an easy means to skip
text without deep analysis.

The majority of existing information systems
are applied to English text. A major drawback of previous systems was their
restrictive degree of portability towards new domains and tasks which
was also caused by a restricted degree of re-usability of the knowledge
sources. Consequently, the major goals which were  identified
during the sixth message understanding conference (MUC-6)
were, on the one hand, to demonstrate task-independent component
technologies of information extraction, and, on the other hand, to
encourage work on increasing portability and ``deeper understanding''
(cf. \cite{GrishmanSundheim:96b}).

In this paper we report on \smes\, 
an information extraction core system for real world
German text processing. The main research topics we are concerned with
include easy portability and adaptability of the core system to extraction
tasks of different complexity and domains. In this paper we will
concentrate on the technical and implementational aspects of the
IE core technology used for achieving the desired portability.
We will only briefly describe some of the current applications built on
top of this core machinery (see section \ref{current-applications}).

\section{The overall architecture of \smes}

The basic design criterion of the \smes\ system is to provide a set
of basic powerful, robust, and efficient natural language
components and generic linguistic knowledge sources
which can easily be customized for processing different tasks in a 
flexible manner. 
Hence, we view \smes\ as a {\em core information extraction system}. 
Customization is achieved in the following directions:

\begin{itemize}
\item   defining the flow of control between modules (e.g., cascaded
        and/or interleaved)
\item   selection of the linguistic knowledge sources
\item   specifying domain specific knowledge
\item   defining task-specific additional functionality
\end{itemize}

\begin{figure}[ht]
\begin{center}~
\setlength{\unitlength}{0.006875in}%
\begin{picture}(462,580)(180,245)
\thicklines
\put(280,340){\framebox(140,60){}}
\put(280,700){\framebox(140,60){}}
\put(280,580){\framebox(140,60){}}
\put(280,460){\framebox(140,60){}}
\put(340,800){\vector( 0,-1){ 40}}
\put(340,700){\vector( 0,-1){ 60}}
\put(340,580){\vector( 0,-1){ 60}}
\put(340,460){\vector( 0,-1){ 60}}
\put(340,340){\vector( 0,-1){ 80}}
\put(180,460){\framebox(60,140){}}
\put(211,600){\vector( 1, 2){ 66.200}}
\put(241,560){\vector( 3, 4){ 39}}
\put(241,511){\vector( 2,-1){ 40}}
\put(210,459){\vector( 3,-4){ 67.920}}
\put(480,481){\framebox(60,140){}}
\put(341,299){\line( 1, 0){172}}
\put(513,299){\vector( 0, 1){181}}
\put(561,601){\framebox(81,79){}}
\put(560,650){\line(-1, 0){ 31}}
\put(529,650){\vector( 0,-1){ 30}}
\put(541,579){\line( 1, 0){ 59}}
\put(600,579){\vector( 0, 1){ 22}}
\put(300,730){\makebox(0,0)[lb]{\raisebox{0pt}[0pt][0pt]{\tiny Tokenizer}}}
\put(300,610){\makebox(0,0)[lb]{\raisebox{0pt}[0pt][0pt]{\tiny Morph./Lex.}}}
\put(300,592){\makebox(0,0)[lb]{\raisebox{0pt}[0pt][0pt]{\tiny Processing}}}
\put(300,495){\makebox(0,0)[lb]{\raisebox{0pt}[0pt][0pt]{\tiny Fragment}}}
\put(300,477){\makebox(0,0)[lb]{\raisebox{0pt}[0pt][0pt]{\tiny Processing}}}
\put(285,375){\makebox(0,0)[lb]{\raisebox{0pt}[0pt][0pt]{\tiny Fragment 
comb.}}}
\put(285,357){\makebox(0,0)[lb]{\raisebox{0pt}[0pt][0pt]{\tiny Template gen.}}}
\put(320,810){\makebox(0,0)[lb]{\raisebox{0pt}[0pt][0pt]{\tiny ASCII Text}}}
\put(320,245){\makebox(0,0)[lb]{\raisebox{0pt}[0pt][0pt]{\tiny Templates}}}
\put(385,310){\makebox(0,0)[lb]{\raisebox{0pt}[0pt][0pt]{\tiny Marked-up 
Text}}}
\put(184,540){\makebox(0,0)[lb]{\raisebox{0pt}[0pt][0pt]{\tiny K-Base}}}
\put(486,561){\makebox(0,0)[lb]{\raisebox{0pt}[0pt][0pt]{\tiny HTML}}}
\put(486,540){\makebox(0,0)[lb]{\raisebox{0pt}[0pt][0pt]{\tiny Interface}}}
\put(580,655){\makebox(0,0)[lb]{\raisebox{0pt}[0pt][0pt]{\tiny Netscape}}}
\put(581,634){\makebox(0,0)[lb]{\raisebox{0pt}[0pt][0pt]{\tiny Browser}}}
\end{picture}

\end{center}
        \caption{A blueprint of the core system}
        \label{IE-archi}
\end{figure}
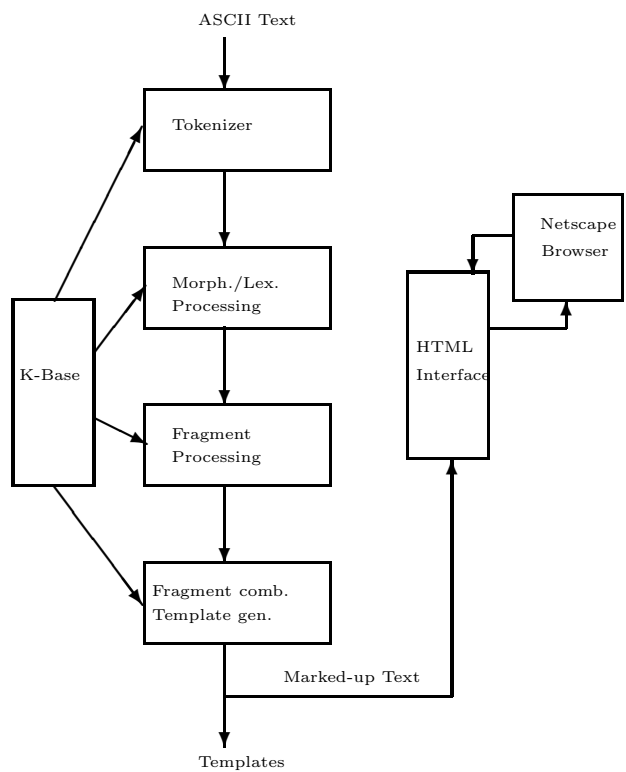

Figure \ref{IE-archi} shows a blueprint of the core system (which
roughly follows the design criteria of
the generic information extraction system described in \cite{Hobbs:92}) .
The main components are: 

  A tokenizer based on regular expressions:
        it scans an ASCII text file for recognizing
        text structure, special tokens like
        date and time expressions, abbreviations and words.

  A very efficient and robust German morphological
        component which performs morphological inflection and compound
        processing. For each analyzed word it returns a (set of)
        triple containing the stem (or a list of stems in case of a
        compound), the part of speech, and inflectional information.
        Disambiguation of the morphological output is performed 
        by a set of word-case sensitive rules, 
        and a Brill-based unsupervised tagger.

  A declarative specification tool for expressing 
        finite state grammars for
        handling word groups and phrasal entities (e.g., general 
        NPs, PPs, or verb groups, 
        complex time and date expressions, proper name
        expressions). A finite state grammar consists of a set of
        fragment extraction patterns defined as finite state transducers
        (FST), where modularity is achieved through  a generic
        input/output device. FST are compiled to
        Lisp functions using an extended version of the compiler
        defined in \cite{Krieger:87}.

   A bidirectional lexical-driven shallow parser for the
        combination of extracted fragments. 
        Shallow parsing is basically directed through
        {\em fragment combination patterns} FCP of the form \tuple{FST_{left},
        anchor, FST_{right}}, where $anchor$ is a lexical entry (e.g., a
        verb like ``to meet'') or a name of a class of lexical entries 
        (e.g., ``transitive-verb''). FCPs are
        attached to lexical entries (e.g., verbs), and are
        selected right after a corresponding lexical entry has been
        identified. They are applied to
        their left and right stream of tokens of recognized fragments.
        The fragment combiner is used for recognizing and extracting
        clause level expressions, as well as for the instantiation of 
        templates.

   An interface to TDL, 
        a type description language for constraint-based grammars 
        \cite{Krieger&Schaefer:COLING-94}.
        TDL is used in \smes\ for performing type-driven
        lexical retrieval, e.g., for concept-driven filtering, 
        and for the evaluation of syntactic agreement tests during
        fragment processing and combination.

   The knowledge base is the collection of different knowledge
        sources, viz. lexicon, subgrammars, clause-level expressions,
        and template patterns. Currently it includes 120.000 
        lexical root entries,
        subgrammars for simple and complex date and time expressions,
        person names, company names, currency expressions, as well as shallow 
        grammars for general nominal phrases, prepositional phrases,
        and general verb-modifier expressions.

Additionally to the above mentioned components there also exists a
generic graphical editor for text items and an HTML interface to the
Netscape browser which performs marking of the relevant text parts by
providing typed parentheses which also serve as links to the internal
representation of the extracted information.

There are two important properties of the system for supporting
portability:
\begin{itemize}
\item   Each component outputs the resulting structures uniformly
        as feature value structures, together with its type and
        the corresponding
        start and end positions of the spanned input expressions.
        We call these output structures {\em text items}.

\item   All (un-filtered) resulting structures of each component are cached
        so that a component can take into account
        results of all previous components. This allows for the definition
        of cascaded as well as interleaved flow of control. 
        The former case means that it is possible to apply
        a cascade of finite state expressions (comparable to that
        proposed in \cite{Fastus:93}), and the latter supports
        the definition of finite state expressions which incrementally
        perform a mix of  keyword spotting, fragment processing, and
        template instantiation.\footnote{Of course, it is also 
        possible---and usually the case in our current applications---to 
        combine both sorts of control flow.}

\end{itemize}

The system has already successfully been applied to classifying
event announcements made via email, scheduling of meetings also sent via
email, and extraction of company information from on-line newswires
(see \ref{current-applications} for more details).
In the next section, we are describing some of the components' properties
in more detail.

\section{Word level processing}

\paragraph{Text scanning}
Each file is firstly preprocessed by the {\it text scanner}. Applying
regular expressions (the text scanner is implemented in lex, the
well-known Unix tool), the text
scanner identifies some text structure (e.g., paragraphs,
indentations), word, number, date and time tokens (e.g, ``1.3.96'',
``12:00 h''), and expands abbreviations.
The output of the text scanner is a stream of tokens, where each word
is simply represented as a string of alphabetic characters (including
delimiters, e.g. ``Daimler-Benz''). Number, date and time expressions are
normalized and represented as attribute values structures. For example
the character stream ``1.3.96'' is represented as
{\sf (:date ((:day 1)(:mon 3)(:year 96))}, and ``13:15 h'' as 
{\sf (:time ((:hour 13)(:min 15)))}.

\paragraph{Morphological processing}
follows text scanning and performs
inflection, and processing of compounds. The capability of efficiently
processing compounds  is crucial since compounding is a very
productive process of the German language.

The morphological component called \mona\  is a descendant of
\morphix, a fast
classification-based morphology  component for German
\cite{Morphix:88}. \mona\ improves \morphix\ in that the
classification-based approach has been combined with the well-known
two-level approach, originally developed by
\cite{Koskenniemi83}. Actually, the extensions concern

\begin{itemize}

\item   the use of tries (see \cite{AHU:83}) as the sole storage
        device for all sorts of lexical information in \mona\
        (e.g., for lexical entries, prefix, inflectional endings), 
        and

\item   the analysis of compound expressions which is realized by means of
        a recursive trie traversal. During traversal two-level rules
        are applied for recognizing linguistically well-formed 
        decompositions of the word form in question.

\end{itemize}

The output of \mona\ is the word form together with all its readings. A
reading is a triple of the form \tuple{stem,inflection,pos}, where
$stem$ is a string or a list of strings (in the case of compounds),
$inflection$ is the inflectional information, and $pos$ is the part of
speech.

Currently, \mona\ is used for the German and Italian language. 
The German version has a very broad coverage (a lexicon of
more then 120.000 stem entries), and an excellent speed (5000 words/sec
without compound handling, 2800 words/sec with compound processing
(where for each compound all lexically possible
decompositions are computed).\footnote{
Measurement has been performed on a Sun~20 using an on-line lexicon of
120.000 entries.}

\paragraph{Part-of-speech disambiguation}
Morphological ambiguous readings are disambiguated wrt. part-of-speech
using case-sensitive rules\footnote{
Generally, only nouns (and proper names) are written in standard German
with an capitalized initial letter  (e.g., ``der Wagen'' {\em the car}
vs. ``wir wagen'' {\em we venture}).
Since typing errors are relatively rare in press releases (or similar
documents) the application of case-sensitive rules are a reliable and
straightforward tagging means for the German language.
} 
and filtering rules which have
been determined using Brill's un-supervised tagger \cite{Brill:95}.
The filtering rules are also used for tagging unknown words.

The filtering rules are determined on the basis of unannotated
corpora. Starting from untagged
corpora, \mona\ is used for initial tagging, where unknown words are
ambiguously  tagged as noun, verb, and adjective. 
Then, using contextual information from
unambiguously analysed word forms, filter rules are determined which
are of the form {\em change tag of word form from noun or verb to noun if the
previous word is a determiner}.

First experiments using a training set of 100.000 words and 
a set of about 280 learned filter rules yields a tagging accuracy 
(including tagging of unknown words) of 91.4\%.\footnote{
Brill reports a 96\% accuracy using a training set
of 350.000 words and 1729 rules. However, he
does not handle unknown words. In \cite{AoneHausman:96}, an extended
version of Brill's tagger is used for tagging Spanish texts, which
includes unknown words. They report an accuracy of 92.1\%.}

Note that the un-supervised tagger required no hand-tagged corpora and
considered unknown words. We expect to increase the accuracy 
by improving the un-supervised tagger through the use of more
linguistic information determined by \mona\, 
especially for the case of unknowns words.

\section{Fragment processing}

Word group recognition and extraction is performed through fragment
extraction patterns which are expressed as finite state transducers
(FST) and which are compiled to Lisp functions using a compiler based
on \cite{Krieger:87}. An FST
consists of a unique name, the recognition part, the output description, 
and a set of compiler parameters. 

\paragraph{The recognition part}

An FST operates on a stream of tokens. The recognition part of an FST
is used for describing regular patterns over such token streams. 
For supporting modularity the different possible kind of tokens are
handled via {\em basic edges}, where a basic edge can be viewed as a
predicate for a specific class of tokens. More precisely a basic edge
is a tuple of the form \tuple{name, test, variable}, where $name$ is
the name of the edge, $test$ is a predicate, and $variable$ holds the
current token $T_c$ , if $test$ applied on $T_c$ holds.
For example the following basic edge \mbox{{\sf (:mona-cat  ``partikel'' pre)}}
tests whether $T_c$ produced by \mona\ is a particle, 
and if so binds
the token to the variable $pre$ (more precisely, each variable of a
basic edge denotes a stack, so that the current token is actually
pushed onto the stack).

We assume that for each component of the system for which fragment
extraction patterns are to be defined, a set of basic edges exists. 
Furthermore, we assume that such a set of basic edges remains fix at
some point in the development of the system and thus can be re-used as
pre-specified basic building blocks to a grammar writer.

Using basic edges the recognition part of an FST is then defined as a regular
expression using a functional notation. For example the recognition
part for simple nominal phrases might be defined as follows:
\vspace{1mm}
{\sf

\noindent\hspace{4mm}(:conc
\par\noindent\hspace{8mm}(:star$\leq$n (:mona-cat ``det'' det) 1)
\par\noindent\hspace{8mm}(:star (:mona-cat ``adj'' adj))
\par\noindent\hspace{8mm}(:mona-cat ``n'' noun))
}

\vspace{1mm}

Thus defined, a nominal phrase is the concatenation of  one optional
determiner (expressed by the loop operator :star$\leq$n, where n starts 
from 0 and ends by 1), followed by zero or more adjectives followed by
a noun.

\paragraph{Output description part}
The output structure of an FST is constructed by collecting together the
variables of the recognition part's basic edges followed by some
specific construction handlers.
In order to support re-usability of 
FST to other applications, it is important to separate the construction
handlers from the FST definition. Therefore, the output description
part is realized through a function called {\sc build-item} which
receives as input the edge variables and a symbol denoting the class of the
FST. For example, if :np is used as a type name for nominal
phrases then the output description of the above NP-recognition part is

\begin{center}
        {\sf (build-item :type :np :out (list det adj noun))}.
\end{center}

The function {\sc build-item} then discriminates according to the
specified type and constructs the desired output to some pre-defined
requests (note, that in the above case the variables {\sc det} and {\sc
adj} might have received no token. In that case their default value NIL
is used as an indication of this fact). Using this mechanism it is
possible to define or re-define the output structure without changing
the whole FST.

\paragraph{Special edges}
There exist some special basic edges namely \mbox{{\sf (:var var)}},
\mbox{{\sf (:current-pos pos)}}
and \mbox{{\sf (:seek name var)}}. The edge \mbox{{\sf (:var var)}} is
used for simply skipping or consuming a token without any checks.
The edge {\sf :current-pos}
is used for storing the position of the current token in the variable
{\sf pos}, and the edge {\sf :seek} is used for calling the FST
named {\sf name}, where {\sf var} is used as a storage for the
output of {\sf name}. This is similar to the :seek edge known from
Augmented Transition Networks with the notably distinction that in our
system recursive calls are {\em disallowed}. Thus {\sf :seek} can also
be seen as a macro expanding operator.

The :seek mechanism is very useful in defining modular grammars, since
it allows for a hierarchical definition of finite state grammars, from
general to specific constructions (or vice versa). 
The following example demonstrates the use of these special edges:
\vspace{1mm}
{\sf

\noindent(compile-regexp
\par\noindent\hspace{2mm}(:conc
\par\noindent\hspace{4mm}   (:current-pos start)
\par\noindent\hspace{4mm}   (:alt
\par\noindent\hspace{6mm}    (:seek time-phase time)
\par\noindent\hspace{6mm} (:conc
\par\noindent\hspace{8mm}        (:star$\leq$n (:seek time-expr-vorfield 
vorfield) 1)
\par\noindent\hspace{8mm}     (:seek mona-time time)))
\par\noindent\hspace{4mm}    (:current-pos end))
\par\noindent\hspace{2mm} :name time-expr
\par\noindent\hspace{2mm} :output-desc
\par\noindent\hspace{4mm}  (build-item :type time-expr :start start  
\par\noindent\hspace{20mm}             :end end :out (list vorfield time))))

}
\vspace{1mm}

This FST recognizes expressions like ``sp\"atestens um 14:00 h'' ({\em
by two o'clock at the latest}) with the output description
{\sf ((:out (:time-rel . ``spaet'') (:time-prep . ``um'') (:minute . 0)
   (:hour . 14))
  (:end . 4) (:start . 0) (:type . time-expr))}

\paragraph{Interface to TDL}

The interface to TDL, a typed feature-based language and inference system
is also realized through basic edges. TDL allows the user to define
hierarchically-ordered types consisting of type constraints and feature
constraints, and has been originally developed for supporting
high-level competence grammar development.

In \smes\ we are using TDL for two purposes:
\begin{enumerate}
\item   defining domain-specific type lattices
\item   expressing syntactic agreement constraints
\end{enumerate}

The first knowledge is used for performing concept-based lexical
retrieval (e.g., for extracting word forms which are compatible to
a given super-type, or for filtering out lexical readings which are
incompatible wrt. a given type), and the second knowledge is used for
directing fragment processing and combination, e.g., for filtering out
certain un-grammatical phrases or for extracting phrases of certain
syntactic type.

The integration of TDL and finite state expressions is easily achieved
through the definition of basic edges. For example the edge

\begin{center}
        {\sf (:mona-cat-type (:and ``n'' ``device'') var)}
\end{center}

\noindent will accept a word form which has been analyzed as a noun and whose 
lexical entry type identifier is subsumed by ``device''.
As an example of defining agreement test consider the basic edge
\begin{center}
        {\sf (:mona-cat-unify ``det''\\
                ``[(num \%1)(case \%2 = gen-val) (gender \%3)]''\\
        agr det)}
                              
\end{center}

\noindent which checks whether the current token is a determiner and
whether its inflection information (computed by \mona) 
unifies with the specified constraints (here, it is checked whether 
the determiner has a genitive reading, where
structure sharing is expressed through variables like \%1). If so, agr
is bound to the result of the unifier and token is bound to det.
If in the same FST a similar edge for noun tokens follows which also
makes reference to the variable agr, the new value for agr is
checked with its old value. In this way, agreement information is
propagated through the whole FST.

An important advantage of using TDL in this way is that
it supports the specification of very compact and modular finite expressions. 
However, one might argue
that using TDL in this way could have dramatic effects on the
efficiency of the whole system, if the whole power of TDL would be
used. In some sense this is true. However, in our current system we
only allow the use of type subsumption which is performed by TDL very
efficiently, and constraints used very carefully and restrictively. 
Furthermore, the TDL interface opens up the possibility of integrating
deeper processing components very straightforwardly.

\paragraph{Control parameters}
In order to obtain flexible control mechanisms for the matching phase
it is possible to specify whether an exact match is requested or
whether an FST should already succeed when the recognition part 
matches a prefix of the input string (or suffix, respectively).
The prefix matching mechanism is used in conjunction
with the Kleene :star and the identity edge :var, 
to allow for searching the whole input stream for extracting 
all matching expressions of an FST (e.g., extracting all NP's, 
or time expressions). For example the following FST extracts all genitive NPs
found in the input stream and collects them in a list:
\vspace{1mm}
{\sf

\noindent(compile-regexp
\par\noindent\hspace{2mm} (:star
\par\noindent\hspace{4mm}   (:alt
\par\noindent\hspace{6mm}    (:seek  gen-phrase x)
\par\noindent\hspace{6mm}    (:var dummy)))
\par\noindent\hspace{2mm} :output-desc (build-item :type list :out x  )
\par\noindent\hspace{2mm} :prefix T
\par\noindent\hspace{2mm} :suffix NIL
\par\noindent\hspace{2mm} :name gen-star)
}
\vspace{1mm}

Additionally, a boolean parameter can be used to specify
whether longest or shortest matches should be prefered (the default is
longest match, see also \cite{Fastus:93} where also longest subsuming
phrases are prefered).

\section{Fragment combination and template generation}

\paragraph{Bidirectional shallow parsing}
The combination of extracted fragments is performed by a lexical-driven
bidirectional shallow parser which operates on {\em fragment combination
patterns} FCP which  are attached to lexical entries (mainly verbs).
We call these lexical entries {\em anchors}.

The input stream for the shallow parser consists of a {\em
double-linked}  list of all extracted fragments found in some input text, 
all punctuation tokens and text tokens (like newline or
paragraph) and all found anchors (i.e., all other tokens of the input
text are ignored). The shallow parser then
applies for each anchor its associated FCP. An anchor can be viewed as
splitting the input stream into a left and right input part. Application of
an FCP then starts directly from the input position of the anchor and
searches the left and right input parts for candidate fragments.
Searching stops either if the beginning or the end of a text has been
reached or if some punctuation, text tokens or other anchors 
defined as stop markers have been recognized.

\paragraph{General form of fragment combination patterns}

A FCP consists of a unique name, an recognition part applied on the
left input part and one for the right input part, an output description
part and a set of constraints on the type and number of collected
fragments. As an prototypical case, consider the following
FCP defined for intransitive verbs like {\em to come} or {\em to begin}:
\vspace{1mm}

{\sf
\noindent(compile-anchored-regexp
\par\noindent\hspace{2mm}((:set (cdr (assoc :start ?*)) anchor-pos)
\par\noindent\hspace{4mm}(:set ((:np (1 1) (nom-val (1 1)))) nec)
\par\noindent\hspace{4mm} (:set ((:tmp (0 2))) opt))
\par\noindent\hspace{2mm} ((:dl-list-left
\par\noindent\hspace{4mm}    (:star
\par\noindent\hspace{6mm}     (:alt
\par\noindent\hspace{8mm}      (:ignore-token (``,'' ``;''))
\par\noindent\hspace{8mm}      (:ignore-fragment :type (:time-phase :pp))
\par\noindent\hspace{8mm} (:add-nec (:np :name-np) 
\par\noindent\hspace{20mm}              :np nec lcompl) 
\par\noindent\hspace{8mm}      (:add-opt (:time-expr :date-expr)
\par\noindent\hspace{20mm}                     :tmp opt lcompl))))
\par\noindent\hspace{2mm}   (:dl-list-right
\par\noindent\hspace{4mm}    (:star
\par\noindent\hspace{6mm}     (:alt
\par\noindent\hspace{8mm}      (:ignore-token (``,'' ``;''))
\par\noindent\hspace{8mm}      (:add-nec (:np) :np nec rcompl) 
\par\noindent\hspace{8mm}      (:add-opt (:time-expr :date-expr) 
 \par\noindent\hspace{20mm}                  :tmp opt rcompl)))))
\par\noindent\hspace{2mm} :name intrans
\par\noindent\hspace{2mm} :output-desc (build-item :type :intrans
\par\noindent\hspace{20mm} :out (list anchor-pos lcompl rcompl)))
}
\vspace{1mm}

The first list remembers the position of the active anchor and introduces
two sets of constraints, which are used to
define restrictions on the type and number of necessary and optional
fragments, e.g., the first constraint says that exactly one :np fragment
(expressed by the lower and upper bound in (1 1)) in
nominative case must be collected, where the second constraint says
that at most two optional fragments of type :tmp can be collected.
The two constraints are maintained by the basic edges :add-nec and
:add-opt. :add-nec performs as follows. If the current token is a
fragment of type :np or :name-np then inspect the set named nec and
select the constraint set typed :np . If the current token agrees in
case (which is tested by type subsumption)
then push it to lcompl and reduce the upper bound by 1. Since next time
the upper bound is 0 no more fragments will be considered for the set
nec.\footnote{In some sense this mechanism behaves like 
the subcategorization principle employed in constraint-based lexical 
grammars.} In a similar manner :add-opt is processed.

The edges :ignore-token and :ignore-fragment are used to explicitly
specify what sort of tokens will not be considered by :add-nec or
:add-opt. In other words this means, that each token which is not
mentioned in the FCP will stop the application of the FCP on the current
input part (left or right).

\paragraph{Complex verb constructions}
In our current system, FCPs are attached to main verb entries. 
Expressions which contain modal, auxiliary verbs or separated
verb prefixes are handled by lexical rules which are applied after
fragment processing and before shallow processing.
Although this mechanism turned out to be practical enough for our
current applications, we have defined also complex verb group fragments VGF. 
A VGF is applied after fragment processing took place. 
It collects all verb forms used in a sentence, and returns 
the underlying dependency-based structure. 
Such an VGF is then used as a {\em complex anchor} for the
selection of appropriate fragment combination patterns
as described above. The advantage of verb
group fragments is that they help to handle  more complex constructions
(e.g., time or speech act) in a more systematic 
(but still shallow) way.

\paragraph{Template generation}
An FCP expresses restrictions on the set of candidate fragments to be
collected by the anchor. If successful the set of found fragments
together with the anchor builds up an instantiated template or
frame. In general a template is a record-like structure consisting of
features and their values, where each collected fragment and the anchor
builds up a feature/value pair. An FCP also defines which sort
of fragments are necessary or optional for building up the whole
template. FCPs are used for defining linguistically oriented
general head-modifier construction (linguistically based on dependency
theory) and application-specific database entries. 
The ``shallowness'' of the template construction/instantiation
process depends on the weakness of the defined FST of an FCP.

A major drawback of our current approach is that necessary and
optional constraints are defined together in one FCP. For example, if
an FCP is used for defining generic clause expressions, where
complements are defined through necessary constraints and adjuncts
through optional constraints then it has been shown that the
constraints on the adjuncts can change for different applications. Thus
we actually lack some modularity concerning this issue. A better
solution would be to attach optional constraints directly with lexical
entries and to ``splice'' them into an FCP after its selection.

\section{Coverage of knowledge sources}

The lexicon in use contains more than 120.000 stem entries
(concerning morpho-syntactic information).

The time and date subgrammar covers a wide range of expressions
        including nominal, prepositional, and coordinated expressions,
        as well as combined date-time expressions (e.g., ``vom
        19. (8.00 h) bis einschl. 21. Oktober (18.00 h)'' yields:
        {\sf (:pp (from :np (day . 19) (hour . 8) (minute . 0))
        (to :np (day . 21) (month . 10) (hour . 18) (minute . 0)))})

The NP/PP subgrammars cover e.g., coordinate NPs, different forms of
        adjective constructions, genitive expressions, pronouns.
        The output structures reflects the underlying head-modifier
        relations (e.g., `` Die neuartige und vielf\"altige
        Gesellschaft '' yields:
        {\sf (((:sem (:head ``gesellschaft'') 
                (:mods ``neuartig'' ``vielfaeltig'')
        (:quantifier ``d-det''))
        (:agr nom-acc-val) (:end . 6) (:start . 1)
        (:type . :np)))}

30 generic syntactic verb subcategorization frames are defined by fragment
        combination patterns (e.g, for transitive verb frame).
        Currently, these verb frames are handled by the shallow parser
        with no ordering restriction, which is reasonably because
        German is a language with relative free word order. 
        However, in future work we will investigate the integration of 
        shallow linear precedence constraints.

The specification of the current data has been performed on
a tagged corpora of about 250 texts (ranging in size from a third  to one
page) which are about event 
announcement, appointment scheduling and business news following
a bottom-up grammar development approach.

\section{Current applications}
\label{current-applications}
On top of \smes\ three application systems have been implemented:

\begin{enumerate}

\item   appointment scheduling via email:
        extraction of co-operate act, duration, range, appointment,
        sender, receiver, topic
\item   classification of event announcements sent via email:
        extraction of speaker, title, time, and location
\item   extraction of company information from newspaper articles:
        company name, date, turnover, revenue, quality, difference
\end{enumerate}

For these applications the main architecture (as described above),
the scanner, morphology, the set of basic edges,
the subgrammars for time/date and phrasal expressions
could be used basically unchanged. 

In (1) \smes\ is embedded in the  \cosma\ system, a German language
server for existing appointment scheduling agent systems
(see \cite{anlp-cosma}, this volume, for more information).
In case (2) additional FST for the
text structure have been added, since the text structure is an
important source for the location of relevant information.
However, since the form of event announcements is usually not
standardized, shallow NLP mechanisms are necessary. Hence, 
the main strategy realized is a mix of
text structure recognition  and restricted shallow
analysis. For application (3), new subgrammars for company names and currency
expressions have to be defined,
as well as a task-specific reference resolution method.

Processing is very robust and fast (between 1 and 10
CPU seconds (Sun UltraSparc)
depending on the size of the text which ranges from very short texts
(a few sentences) upto short texts (one page)).
In all of the three applications we obtained high coverage and good
results. Because of the lack of comparable
existing IE systems defined for handling German texts in similar
domains and the lack of evaluation standards for the German language
(comparable to that of MUC), we cannot claim that these results are
comparable.

However, we have now started the implementation of a new application 
together with a commercial partner, where a more systematic evaluation of
the system is carried out. Here, {\sc smes} is applied
on a quite different domain, namely news items concerning the German
IFOR mission in former Yugoslavia. Our task is to identify those
messages which are about violations of the peace treaty and to extract the
information about location, aggressor, defender and victims.

The corpus consists of a set of monthly reports (Jan. 1996 to
Aug. 1996) each consisting of
about 25 messages from which 2 to 8 messages 
are about fighting actions. These messages have been hand-tagged
with respect to the relevant information.
Although we are still in the development phase we will briefly 
describe our experience of adapting {\sc smes} to this new domain.
Starting from the assumption
that the core machinery can be used un-changed we first 
measured the coverage of the existing linguistic knowledge sources.
Concerning the above mentioned corpus the lexicon covers about 90\%.
However, from the 10\% of unrecognized words about 70\% are proper names
(which we will handle without a lexicon) and 1.5\% are spelling
errors, so that the lexicon actually covers more then 95\% of this unseen text
corpus. The same ``blind'' test was also carried out for the date,
time, and location subgrammar, i.e., they have been run on the new
corpus without any adaption to the specific domain knowledge. For the
date-/time expressions we obtained a recall of 77\% and a precision
of 88\%, and for the location expressions we obtained 66\% and 87\%,
respectively. In the latter case, most of the unrecognized expressions
concern expressions like ``nach Taszar/Ungarn'', ``im serbischen
bzw. kroatischen Teil Bosniens'', or ``in der Moslemisch-kroatischen
F\"oderation''. For the general NP and PP subgrammars we obtained a
recall of 55\% and a precision of 60\% (concerning correct head-modifier
structure). The small recall is due to some lexical gap (including
proper names) and unforeseen complex expressions like ``die Mehrzahl
der auf 140.000 gesch\"atzten moslemischen Fl\"uchtlinge''. But note
that these grammars have been written on the basis of different corpora.

In order to measure the coverage of the fragment combination patterns FCP,
the relevant main verbs of the tagged corpora have been associated with
the corresponding FCP (e.g., the FCP for transitive verbs),
without changing the original definition of the FCPs. 
The only major change to be done concerned the extension of 
the output description function {\sc build-item} for
building up the new template structure.
After a first trial run we obtained an unsatisfactory recognition rate of about
25\%. One major problem we identified was the frequent use of passive
constructions which the shallow parser was not able to process.
Consequently, as a first actual extension of {\sc smes} to the new domain we
extended the shallow parser to cope with passive constructions.
Using this extension we obtained an recognition of about 40\% after a
new trial run. 

After the analysis of the (partially) unrecognized messages (including
the misclassified ones), we identified the following major
bottlenecks of our current system. First, many of the partially
recognized templates are part of coordinations (including
enumerations), in which case several (local) templates share the same
slot, however this slot is only mentioned one time. Resolving this kind
of ``slot sharing'' requires processing of elliptic expressions of
different kinds as well as the need of domain-specific inference
rules which we have not yet foreseen as part of the core system.
Second, the wrong recognition of messages
is often due to the lack of semantic constraints which would be applied
during shallow parsing in a similar way as the subcategorization
constraints. 

Although these current results should and can be improved we are
convinced that the idea of developing a core IE-engine is a
worthwhile venture.

\section{Related work}
In Germany, IE based on innovative language technology is still
a novelty. The only groups which we are aware of which also consider
NLP-based IE are \cite{Hahn:92,Bayer:94}.
None of them make use of such sophisticated components, as we do in \smes.
Our work is mostly influence by the work of 
\cite{Hobbs:92,Fastus:93,Grishman:96} as well as by the work described
in \cite{Jasper:92,Gemini:93}.

\section{Conclusion}
We have described an information extraction core system for real world 
German text processing. The basic design criterion of the 
system is of providing a set of basic powerful, robust, 
and efficient natural language components and generic linguistic 
knowledge sources which can easily be customized for processing 
different tasks in a  flexible manner. 
The main features are: a very efficient and robust
morphological component, a powerful tool for expressing finite state
expressions, a flexible bidirectional shallow parser, as well as
a flexible interface to an advanced formalism for typed
feature formalisms. 
The system has been fully implemented in Common Lisp and C.

Future research will focus towards automatic adaption and acquisition methods,
e.g., automatic extraction of subgrammars from a competence base
and learning methods for domain-specific extraction patterns.
\label{future-work}

\section{Acknowledgement}
The research underlying this paper was
supported by research grants from the German
Bundesministerium f\"ur Bildung, Wissenschaft, Forschung und
Technologie (BMBF) to the
DFKI projects {\sc paradice}, FKZ~ITW~9403 and {\sc paradime}, 
FKZ~ITW~9704.
We would like to thank the following people for fruitful discussions:
Hans Uszkoreit, Gregor Erbach, and Luca Dini.

\bibliographystyle{acl}
{\small

}

\end{document}